\documentclass[10pt,twocolumn,letterpaper]{article}

\usepackage{ijcb}
\usepackage{times}
\usepackage{epsfig}
\usepackage{graphicx}
\usepackage[table,xcdraw]{xcolor}
\usepackage{amsmath}
\usepackage{amssymb}
\usepackage{caption}
\usepackage{enumitem}
\usepackage[norule,symbol,perpage]{footmisc}
\usepackage{multirow}

\ijcbfinalcopy 

\ifijcbfinal\fi

\makeatletter
\def\ps@IEEEtitlepagestyle{
\def\@oddfoot{\mycopyrightnotice}
\def\@evenfoot{}
}
\def\mycopyrightnotice{
{\hfill \footnotesize 978-1-7281-9186-7/20/\$31.00 \copyright 2020 IEEE\hfill}
}
\makeatother

\begin{document}

\title{On Benchmarking Iris Recognition within a Head-mounted Display for AR/VR Applications}

\author{Fadi Boutros$^{1,2}$, Naser Damer$^{1,2}$, Kiran Raja$^{3,4}$, Raghavendra Ramachandra$^{4}$,
\\
Florian Kirchbuchner$^{1,2}$, Arjan Kuijper$^{1,2}$\\
$^{1}$Fraunhofer Institute for Computer Graphics Research IGD,
Darmstadt, Germany\\
$^{2}$Mathematical and Applied Visual Computing, TU Darmstadt,
Darmstadt, Germany\\
$^{3}$ The Norwegian Colour and Visual Computing Laboratory, NTNU, Gjovik, Norway\\
$^{4}$ Norwegian Biometrics Laboratory, NTNU, Gjovik, Norway\\
 Email: {fadi.boutros@igd.fraunhofer.de}
}


\maketitle
\thispagestyle{empty}

\begin{abstract}
Augmented and virtual reality is being deployed in different fields of applications.
Such applications might involve accessing or processing critical and sensitive information, which requires strict and continuous access control. Given that Head-Mounted Displays (HMD) developed for such applications commonly contains internal cameras for gaze tracking purposes, we evaluate the suitability of such setup for verifying the users through iris recognition. In this work, we first evaluate a set of iris recognition algorithms suitable for HMD devices by investigating three well-established handcrafted feature extraction approaches,
and to complement it, we also present the analysis using four deep learning models.  While taking into consideration the minimalistic hardware requirements of stand-alone HMD, we employ and adapt a recently developed miniature segmentation model (EyeMMS) for segmenting the iris. Further, to account for non-ideal and non-collaborative capture of iris, we define a new iris quality metric that we termed as Iris Mask Ratio (IMR)
to quantify the iris recognition performance. Motivated by the performance of iris recognition, we also
propose the continuous authentication of users in a non-collaborative capture setting in HMD. Through the experiments on a publicly available OpenEDS dataset, we show that performance with $EER = 5\%$ can be achieved using deep learning methods in a general setting, along with high accuracy for continuous user authentication.
\end{abstract}

\let\thefootnote\relax\footnotetext{\mycopyrightnotice}

\vspace{-2mm}
\section{Introduction}
\vspace{-1mm}

The commonly implemented security mechanisms in HMD devices depend on pattern matching or Personal Identification Number (PIN)~\cite{george2017seamless}, which is limited to individual knowledge. An alternative for PIN and pattern based passwords, earlier works have proposed to use biometric data to authenticate the Virtual Reality and Augmented Reality (VR/AR) users based on their natural interactions within the virtual space~\cite{kupin2019task}. However, this approach is limited to specific application scenarios where the user performs a defined physical task. Further, the VR/AR applications, such as field policing and crime scene investigation~\cite{poelman2012if}, may require accessing or transmitting sensitive information. A user in such a scenario should be properly and continuously authenticated to prevent anonymous access to sensitive data and to guarantee the safe use of the system in multi-user environments.
These, among many other application scenarios, rise a question regarding the security mechanism in such headset devices.

Noting that HMD for AR/VR applications is commonly built with internal cameras to enable gaze interaction with the virtual environment, our assertion is that such a camera can be used to verify the user's identity based on iris patterns, even in a continuous manner. As simplistic idea may appear, in such a verification scenario, and in comparison to the traditional iris recognition applications, we observe two main challenges.
The first one is related to the limited computational and storage power of HMD devices. The current HMD device in the market is supplied with mobile processor with up to 4 GB of memory. For instance, the recent device from HTC Vive has a Qualcomm Snapdragon 835 processor with 4GB of memory. Under such hardware constraints, large segmentation methods or feature extraction models are not realistically deployable, especially when considering parallel deployment with other applications. The second challenge concerns the non-cooperative nature of the image capture process where the eye image is captured without user cooperation at high frame rate to enable seamless interaction with VR/AR application \cite{DBLP:journals/corr/abs-1905-03702}. Common iris recognition systems require significant cooperation of the user to capture high quality iris image with widely opened eyes. While in the AR/VR scenario, the user should not be required to continuously and intentionally collaborate with the identity verification sub-system impeding the use of AR/VR, rather the system should run in the background. Such an interaction result in sub-optimal iris captures unlike the traditional iris recognition systems leading to performance degradation. 

In this paper, we investigate the possibility of using iris images captured from internal cameras of HMD for user verification. Considering the hardware constraints of HMD devices, we first utilize and adapt a recently developed miniature segmentation model \cite{Boutros_2019_ICCV}. Further, to provide the evaluation in a holistic manner, we explore three well-used handcrafted feature extraction methods and four deep learning models for iris recognition. 
Considering the challenges in adapting the iris recognition directly for HMD data, we propose a new iris quality metric - Iris Mask Ratio (IMR) to suitable select the iris image for verification purposes. Further, we propose a continuous authentication model using iris recognition for verifying the users of HMD devices. 

\vspace{-2mm}
\section{Related works on HMD}
\label{sec:related-works}
\vspace{-1mm}
One of the most accurate and widely deployed approaches to extract iris features is inspired by the method proposed by Daugman~\cite{daugman2009iris}. Further, iris recognition approaches have been proposed in the literature, whether they are derivative of Daugman’s iris features or based on deep learning techniques.  Sun and Tan~\cite{sun2008ordinal} presented ordinal measures (OM) as a novel iris features. Damer et al. \cite{DBLP:journals/spl/DamerTBK17} proposed the transformation of iris features to a rotation invariant space. K. Miyazawa~\cite{miyazawa2008effective} proposed an approach based on Discrete Fourier Transforms (DFT). More recently, an approach was presented by Chen J.et al.~\cite{chen2016iris}, where they build a new set of iris features based on Human-interpreted Crypts Features.

Recently, several works have explored the use of deep learning techniques for iris recognition. Liu et al \cite{DBLP:journals/prl/LiuZLST16} proposed DeepIris network to learn pairwise filters and the deep representations of heterogeneous iris images.
Gangwar and Joshi \cite{DBLP:conf/icip/GangwarJ16} proposed the DeepIrisNet network achieving superior performance on multiple datasets \cite{DBLP:journals/pami/PhillipsSOFBSS10} datasets.  Nguyen et al. \cite{DBLP:journals/access/NguyenFRS18} investigated the performance of several pre-trained CNNs (on ImageNet) for iris recognition. 
Similarly, Zanlorensi \cite{DBLP:journals/iet-bmt/ZanlorensiLJPM20} fine-tuned ResNet and VGG models
for cross-spectral ocular recognition.

Deep learning approach has been used for iris segmentation. Some of approaches are based on U-Net \cite{lozej2018end}, Fully Convolutional Network (FCN) \cite{liu2016accurate}, Context Encoding Network \cite{chen2018encoder}, Cascade Refinement Network (CRN) \cite{Boutros_2019_ICCV}, Encoder-Decoder \cite{Perry_2019_ICCV}, Fast-SCNN  \cite{DBLP:journals/corr/abs-1902-04502}
and SegNet \cite{rot2018deep}. A few of these works addressed efficient segmentation e.g. for embedded device. 
Among the listed models,
the Eye-MMS model proposed Boutros et al. \cite{Boutros_2019_ICCV} and MinENet proposed by Perry and Fernandez \cite{Perry_2019_ICCV} are the smallest models with 80K and 222K  trainable parameters, respectively.

Along the lines of verifying the HMD user, a recently H2020 EU-funded project is investigating the use of AR headsets for border guards, in both crowded border-crossing points and remote locations 
\cite{ARESIBO}. Such application scenarios need users to be verified in a continuous manner, and without disturbance where the information processed and displayed are of secured nature. 
Lee et al. \cite{DBLP:conf/icba/LeeNPK04} designed a prototype of gaze estimation and iris recognition using a camera attached inside a wearable headset. 
However, this approach required user cooperation as the experimental setup assumed that the user should look directly to the camera when  authentication is needed. Although the reported  verification performance was very promising, the performed experiments are different from the real HMD scenario where the eye is captured on the fly without user cooperation at a high speed frame rate (around 200Hz \cite{DBLP:journals/corr/abs-1905-03702}). 

A recent work by Bastias et al.~\cite{bastias2017method} proposed a method for iris reconstruction from several 2D near-infrared iris images from a custom sensor for 2D image capturing mounted on a wearable headset. However, the work did not target wearable headsets specifically, rather used it to create a capturing setup and proposed a consequent verification approach. 
Very recently, Boutros et al. \cite{9107939} evaluated the possibility of using images captured from an HMD internal camera for periocular biometric. The work also presented a reference sample selection strategy to enhance the periocular verification performance within HMD environment. 

Addressing the internal capture in HMD, recently, the OpenEDS database was released \cite{DBLP:journals/corr/abs-1905-03702} which is a large scale eye images dataset captured using a virtual-reality HMD device with two eye-facing cameras. Based on the OpenEDS, Facebook hosted competitions for two main challenges, semantic segmentation, and synthetic eye generation \cite{facebookC}, targeting gaze-tracking solutions. 
To deal with the absence of a large scale identity-specific images captured from HMD cameras, Damer et al. \cite{9021978} proposed an identity-preserving synthetic ocular image generation model based on OpenEDS that can be used for training propose.
Although the target of OpenEDS is gaze-tracking, it opens an opportunity to evaluate iris recognition in HMD setups forming the basis of our work. Motivated by the limited works investigating iris biometrics in HMD devices, we investigate the suitability of iris recognition by benchmarking six different algorithms. Further, we propose a continuous authentication approach suitable for HMD environment. 
\begin{figure}[t]
\begin{center}
 \includegraphics[width=0.98\linewidth]{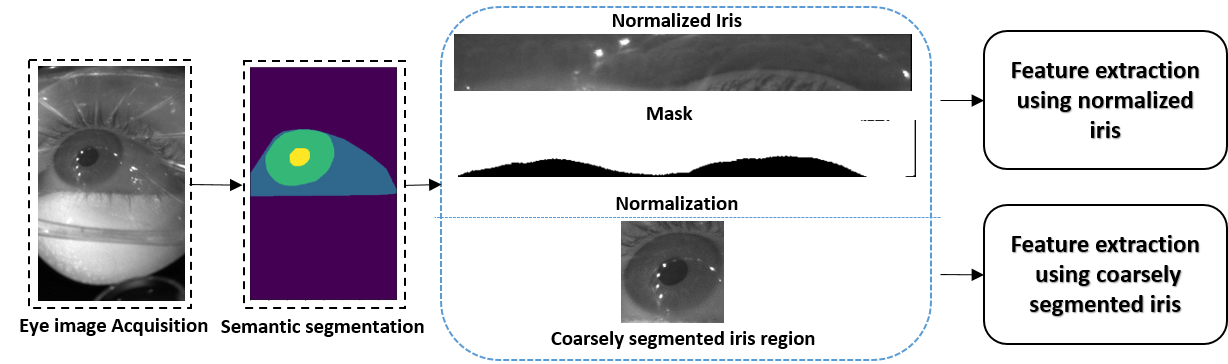}
\end{center}
  \vspace{-4mm}
   \caption{Overview of iris recognition workflow in HMD. 
   }
   \vspace{-4mm}
\label{fig:workfloa}
\end{figure}

\section{Traditional iris recognition in HMD}
\label{sec:iris-recognition-hmd}
The main goal of this work is to provide a comprehensive study on iris recognition solutions within AR/VR environment.
The generic iris recognition pipeline consists of four main steps: a) image acquisition,  b) image preprocessing, d) feature extraction, e) comparison and decision making.  
These steps are  illustrated in Figure ~\ref{fig:workfloa}. 
The image captured in the HMD contains areas beyond the iris, therefore, we segment the iris region and extract the normalized iris.
For different recognition methods, we utilized the normalized iris or the coarse iris region to extract iris features. 
This section presents the used segmentation model, proposed quality metric, and the different  feature extraction methods.

\subsection{Iris segmentation and normalization}
\label{sec:seg}

HMD captured images contain regions beyond the iris, as shown in Figure~\ref{fig:workfloa}. As we baseline iris recognition using approaches with different requirements, we apply two kinds of segmentation, fine and coarse segmentation.   
Considering the minimalistic computational power of HMD devices, we employ computationally-light weight segmentation. 
As the segmentation should be able to scale-up to images acquired in non-collaborative scenario, we note that Weighted Adaptive Hough and Ellipso-polar Transform \cite{DBLP:conf/icb/UhlW12} or the Contrast-adjusted Hough Transform \cite{masek2003recognition} may not be suitable as the iris images are non-ideal.

Given the requirements stated above, we opted to use the Miniature Multi-scale Segmentation Network (Eye-MMS) approach proposed recently \cite{Boutros_2019_ICCV}. The compact size of the Eye-MMS model with only 80K parameters suits minimalist hardware specifications while achieving high accuracy as demonstrated on the OpenEDS database \cite{DBLP:journals/corr/abs-1905-03702}. In comparison to the one of the latest general light-weight semantic segmentation methods, the Fast-SCNN model \cite{DBLP:journals/corr/abs-1902-04502} contains 1.1 million parameters. The MinENet proposed by Perry and Fernandez \cite{Perry_2019_ICCV} achieved slightly better performance on the OpenEDS database, however, with 222K parameters.

We thus employ Eye-MMS model\cite{Boutros_2019_ICCV} as detailed in the Section~\ref{sec:exp} to segment the iris region by labelling the background, sclera, iris, and pupil areas. To neglect any irregularly labeled pixels, we post-process the semantic segmentation
by fitting a convex hull around the largest found contours around each of the considered labels.
These hulls represent the borders of each label and from the labelled iris region, we extract the iris for normalization. To normalize the iris region efficiently, we start by defining a general circular border that contains the pupil and the iris. The pupil circular region is defined around its center of moment and has the radius of the closest (from the center of the moment) pupil labeled pixel. The circular border between the iris and the sclera is also defined centered around the pupil center of moment and has the radius of the distance between this center and the furthest (from the center of the moment) iris labeled pixel.
The iris is normalized using the rubber sheet model by unrolling it to a rectangular image \cite{daugman2004iris}. Similarly, a mask map is created with zero values for each label not belonging to iris, and ones for each pixel labeled as iris. Alongside the fine segmentation, we also carry out a coarse segmentation by cropping the area of eye image containing only the iris and pupil. To do that, we use the segmentation labels and calculate the rectangular bounding box that contains the pupil and the iris. An example of the segmentation, normalization, masking, and coarsely segmented iris, is presented as part of the workflow in Figure~\ref{fig:workfloa}.

\subsection{Proposed quality metric for iris selection} 
To account for the non-cooperative nature of the iris image acquisition in HMD devices where the iris images are non-ideal, we define a new quality metric for choosing the iris images prior to feature extraction. We introduce a new metric by defining the Iris Mask Ratio (IMR) as a ratio of the actual iris area (mask neglected) size to the whole normalized image size. A higher IMR indicates that a larger proportion of the iris is visible in the image, making it more suitable for feature extraction and comparison. The IMR is used to select the reference image from the reference images pool of each identity, i.e. the image with the highest IMR is selected from each reference pool to be the reference. 

The defined quality metric is also used as a basis for choosing the probe images appropriately, i.e. selecting iris images to be used for verification from the series of iris images. We threshold the IMR value of iris images to neglect images with low IMR. To validate the applicability of the proposed quality metric, we correlate it to verification performance as explained in the experimental section.

\subsection{Iris feature extraction and comparison}

We utilize three well-established and complementary handcrafted iris feature extraction methods owing to the robustness and time-tested applicability for iris recognition in various constrained and unconstrained settings \cite{bowyer2008image, shah2009iris, kumar2010comparison, ortega2009multiscenario, proencca2016unconstrained, ross2010iris}. The iriscodes in the first handcrafted method are extracted using the classical Gabor features as proposed by Daugman \cite{daugman2004iris} and we employ the generalized version of the same by using 1D Log-Gabor features \cite{masek2003recognition}. The second handcrafted approach uses Discrete Cosine Transform (DCT) coefficients of overlapped angular patches from normalized iris images to derive the iriscodes \cite{monro2007dct}. The third handcrafted approach extracts the iriscodes using the Cumulative-Sum-Based Change Analysis \cite{ko2007novel}. For all the three handcrafted feature extraction schemes, we employ Hamming Distance (HD) measure to compute the similarity between two irises. Further, noting the early works pointed the benefits of the Shifted Hamming Distance (SHD)  \cite{rathgeb2010secure, rathgeb2011shifting} to account for rotational and displacement invariance, we use the SHD by shifting the iriscodes by 8 bit in both directions to obtain the scores. The minimum of all the SHDs computed is further used as comparison scores for reporting the performance in this work.

Additionally,  we investigate the performance of four CNN solutions, the DeepIrisNet \cite{DBLP:conf/icip/GangwarJ16}, the MobileNetV3 \cite{howard2019searching}, and transfer learning on ResNet \cite{DBLP:conf/cvpr/HeZRS16} and DenseNet \cite{DBLP:conf/cvpr/HuangLMW17}.
DeepIrisNet \cite{DBLP:conf/icip/GangwarJ16} is a specifically designed for iris recognition and provides two architectures, DeepIrisNet-A and DeepIrisNet-B, both showing excelent performance. 
We choose the simpler architecture, DeepIrisNet-A, for our evaluation. 
In the second and third approach, we employ Off-the-Shelf CNN features by applying transfer learning on ResNet \cite{DBLP:conf/cvpr/HeZRS16} and DenseNet \cite{DBLP:conf/cvpr/HuangLMW17} models pre-trained on the ImageNet \cite{DBLP:journals/cacm/KrizhevskySH17}. 
Our choice of ResNet-50 and DenseNet-201 was based on their promising reported accuracy for iris recognition \cite{DBLP:journals/access/NguyenFRS18,DBLP:journals/iet-bmt/ZanlorensiLJPM20}.
In the fourth approach, we learn the features using MobileNetV3 \cite{howard2019searching}, which is designed for low resource application. MobileNetV3 considers several optimizations in addition to architecture search to create two networks, MobileNetV3-Large and MobileNetV3-Small. 
The networks are designed for high and low resource application and achieved a higher classification accuracy than previous lightweight models such as MobileNet series \cite{DBLP:journals/corr/HowardZCKWWAA17,DBLP:journals/corr/abs-1801-04381}  and Mnasnet \cite{DBLP:conf/cvpr/TanCPVSHL19}. We employ MobileNetV3-Small architecture due to its fewer parameters (3 million).
We train and evaluate the CNN models on the normalized iris to be compatible with earlier works on deep iris recognition. 
We additionally evaluate the deep representation extracted from coarsely segmented iris region as considered in unconstrained environment by some works \cite{zanlorensi2018impact,DBLP:journals/iet-bmt/ZanlorensiLJPM20}, especially when the accurate segmentation is challenging.

For all models, we modified the number of classes in the classification layer to the number of identities in our training set (95 identities). 
To adapt DeepIrisNet and MobileNet for iris images from HMD devices, we trained these models from scratch on training data of OpenEDs database \cite{DBLP:journals/corr/abs-1905-03702} 
with softmax classifier. Besides, we fine-tuned the entire DenseNet-201 and ResNet-50 models on the same training data \cite{DBLP:journals/corr/abs-1905-03702}.
During testing, 
for each model, 
the softmax layer is removed and features are extracted $f$ from the last layer
which is of dimension $1 \times 1 \times 4096$ in DeepIrisNet, $7 \times 7 \times 2048$ in ResNet-50, $7 \times 7 \times 1920$ in DenseNet-201 and $1 \times 1 \times 1280$ in MobileNet.

\section{Iris continuous authentication in HMD}
\label{sec:continuous-authentication}
Given that the iris images are captured continuously  during the interaction with the HMD, we also propose a framework to continuously authenticate the user employing the captured iris images. We motivate our work based on the trust model for keystroke continuous authentication introduced by Bours et al. \cite{bours2012continuous} and adapted for multi-biometrics in \cite{DBLP:conf/fusion/DamerMB16}. The trust model
measures the confidence of the current user being a genuine or imposter user.
This trust in the genuineness of the user is expressed as a trust value. The adjustment of the trust value depends on the penalty-and-reward function which adjusts the trust value based on the sampling over time/actions.  In HMD, the images are continuously captured for eye tracking. Therefore, the penalty-and-reward function in our model is triggered by each image acquisition. We propose a variable penalty-and-reward function where the trust value is updated based on the distance between the comparison scores of iris and the threshold at the Equal Error Rate (EER). If the score exceeds the threshold, the trust value is increased and the user is treated as the genuine user, vice versa for imposter. The trust value is continuously increased (reward) when the score is higher than the threshold and decreased (penalty) when the score is lower than the threshold or if there is a gap between two captured images (e.g. failing to pass the IMR quality, IMR\_Th). If the trust value falls below the threshold, the system logs the user out.
The trust value is given as:
{\small
\[ TV =
  \begin{cases}
T & \quad \text{at startup} \\
max(TV-\alpha,-1) & \quad \text{if } \text{$IMR$} < \text{$IMR \textunderscore Th$} \\
min (TV+($cs$-T),1) & \quad \text{if } \text{$cs$} \geq \text{$T$} \\
max(TV-(T-$cs$),-1) & \quad \text{if } \text{$cs$} \leq \text{$T$} 
  \end{cases}
\]
}
where $cs$ is the comparison score and $T$ is the EER threshold. At the start of using HMD device, the trust value is set to $T$, where $T$ is the threshold for punishment or reward. 
If there is a sample gap, a penalty of $\alpha=0.01$ (can be adjusted for application) is applied. The lower and upper limits of the trust value are set to $-1$ and $+1$, respectively.

\begin{figure*}[t]
\begin{center}
   \includegraphics[width=0.99\linewidth]{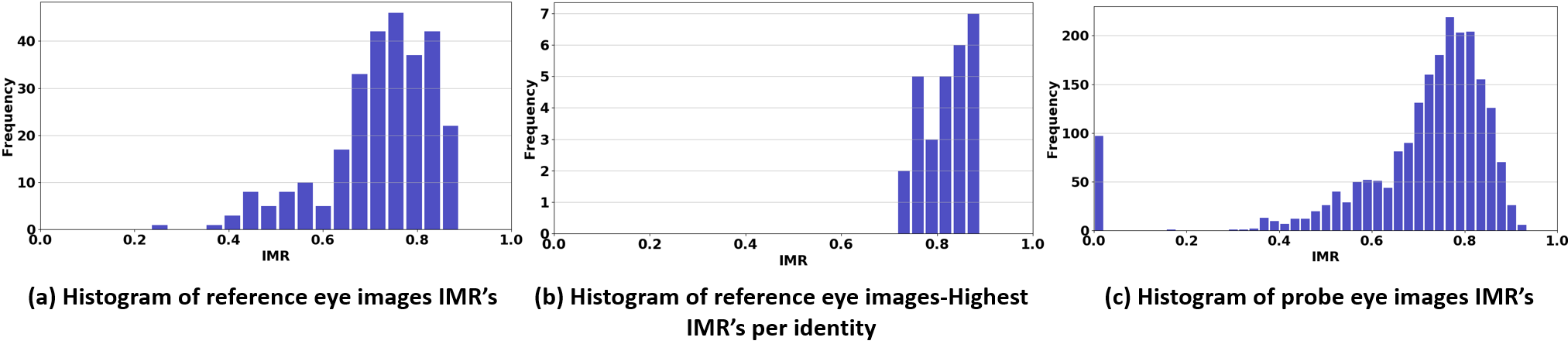}
\end{center}
 \vspace{-7mm}
   \caption{IMR value histogram of the iris images of full reference pool (a), selected references (b), and all probe samples (c). }
\label{fig:imr}
\vspace{-3mm}
\end{figure*}

\section{Experimental setup}
\label{sec:exp}
To evaluate the applicability of iris recognition and the proposed continuous authentication model, we employ OpenEDs database \cite{DBLP:journals/corr/abs-1905-03702}. OpenEDs database is acquired using a virtual-reality HMD with two synchronized eye-facing cameras at a frame rate of 200 Hz under controlled illumination. The semantic segmentation dataset included 12759 images of 152 individuals with a pixel resolution of 640x400. The data is split into 8916 eye images for training, 2403 for validation, and 1440 for testing as described in~\cite{DBLP:journals/corr/abs-1905-03702}. The test split is not available publicly yet and thus, is not used in this work. Since the semantic segmentation labels are available only for both training and validation splits, the segmentation model is trained on training split and evaluated on validation split. 
The deep learning methods are trained on the training split of the semantic segmentation dataset and tested on the validation split. 
We randomly selected a subset of 190 images (two per identity) of the training split to validate the model for early stopping to avoid over-fitting the model during the training.
The normalized iris is reshaped to $128 \times 128$ as proposed in DeepIrisNet approach\footnote{Several input image sizes were evaluated and the highest verification performance was achieved using input size of $128 \times 128$}. For DenseNet, MobileNetV3 and ResNet, the normalized iris are resized to $224 \times 224$ to match their input layer size. Further, the coarsely segmented iris region was resized to $224 \times 224$ to match input layer size of the evaluated models.
As the handcrafted feature extraction approaches do not require training, they are directly evaluated using the validation split. The normalized iris images are resized to a dimension of $512 \times 64$ pixels for extracting the handcrafted features. The validation data is identity-disjoint from the training set.
Each of the 28 validation identities contained between 37 and 128 images captured consecutively and on average 86 per identity. The image set of each identity is split into reference and probe images. The first 10 images for each identity are considered as reference. Further, the initial reference image is chosen by the newly defined quality metric i.e., highest IMR.
The consequent five images are neglected to create a time gap between reference and probe images. All consequent images for each identity are considered as probes. 

\vspace{-2mm}
\paragraph{Segmentation:} We train the Eye-MMS model for 40 epochs with the training parameters as in \cite{Boutros_2019_ICCV}. The results are post-processed as described in Section \ref{sec:seg}. 
The performance of the segmentation model is reported as Intersection over Union (IoU) ratio of each of the four different segmented areas between the ground-truth and predicted label. The unweighted mean IoU of the areas is also reported to provide an overall indication of the performance.

\vspace{-2mm}
\paragraph{Deep learning model training setup:}
The investigated models are trained using Adam optimizer with learning rate of 1e-4 and batch size of 64. 
Each model is trained twice, once using coarsely segmented iris as input and once using normalized iris. We set the initial number of epochs to 100 and the early stopping patience parameter to 5,
causing DeepIrisNet, MobileNetV3, DenseNet, and ResNet to stop after 41, 19, 21 and 29 epochs, respectively on normalized iris training data, and after 23, 19, 23 and 18 epochs, respectively for the coarsely segmented iris images.

\vspace{-2mm}
\paragraph{Iris verification:} We evaluate the verification performance of deep learning methods with cosine-distance for comparison. For each of the deep learning methods, we evaluated representation extracted from coarsely segmented iris region and normalized iris images, resulting in evaluation of eight models. 
For handcrafted feature extraction methods, the LG and the DCT approaches are evaluated with HD and SHD distance for comparison, resulting in four evaluation settings, noted as LG-HD, LG-SHD, DCT-HD, and DCT-SHD. A fifth setup uses the CSBCA features with the HD distance for comparison, as the nature of the feature extraction does not benefit from the computationally more expensive SHD distance. Each of the settings is evaluated with IMR thresholds 0.0 and 0.7 computed on the probes. The verification performance is reported as Receiver Operating Characteristic (ROC) curves, Area under the curve (AUC), False Match Rate (FMR) at fixed False Non-Match Rate (FNMR) (FMR10, the lowest FNMR for FMR$\leq$10\%), and Equal Error Rate (EER).

\begin{table}[!t]
\begin{center}
\resizebox{0.95\linewidth}{!}{
\begin{tabular}{c|c|c|c|c|c|c|}
\cline{2-7}
                              & \textbf{SG 0-1} &\textbf{ SG 2-3} & \textbf{SG 4-5} & \textbf{SG 6-7} & \textbf{SG \textgreater8} &\textbf{ Max SG} \\ \hline
\multicolumn{1}{|c|}{IMR 0}   & 1951  & 0     & 0     & 0     & 0               & 0      \\ \hline
\multicolumn{1}{|c|}{IMR 0.1} & 1856  & 13    & 2     & 7     & 1               & 9      \\ \hline
\multicolumn{1}{|c|}{IMR 0.2} & 1851  & 13    & 2     & 7     & 1               & 9      \\ \hline
\multicolumn{1}{|c|}{IMR 0.3} & 1856  & 13    & 2     & 7     & 1               & 9      \\ \hline
\multicolumn{1}{|c|}{IMR 0.4} & 1850  & 16    & 2     & 7     & 1               & 9      \\ \hline
\multicolumn{1}{|c|}{IMR 0.5} & 1805  & 29    & 5     & 7     & 1               & 10     \\ \hline
\multicolumn{1}{|c|}{IMR 0.6} & 1666  & 53    & 15    & 8     & 4               & 15     \\ \hline
\multicolumn{1}{|c|}{IMR 0.7} & 1397  & 57    & 13    & 16    & 17              & 19     \\ \hline
\end{tabular}%
}
\vspace{-2mm}
\caption{This table provides a view on the size and amount of sequence sample gaps (SG) 
(right most column). Each column represents a certain gap and the numbers of the table represent the occurrences of this gap with a certain IMR threshold setting e.g. SG 0-1 is the case where two consecutive captures do not have any neglected capture or one neglected capture between them.
With a higher IMR threshold, the higher sample gap occurs more often.
}
\label{tab:gap}
\vspace{-5mm}
\end{center}
\end{table}

\section{Results}
\label{sec:results}
In this section, we present the results of the iris segmentation, 
followed by a view on iris image selection by the proposed quality metric. We then discuss iris verification followed by the results of the continuous authentication.

\begin{figure*}[t]
\begin{center}
   \includegraphics[width=0.99\linewidth]{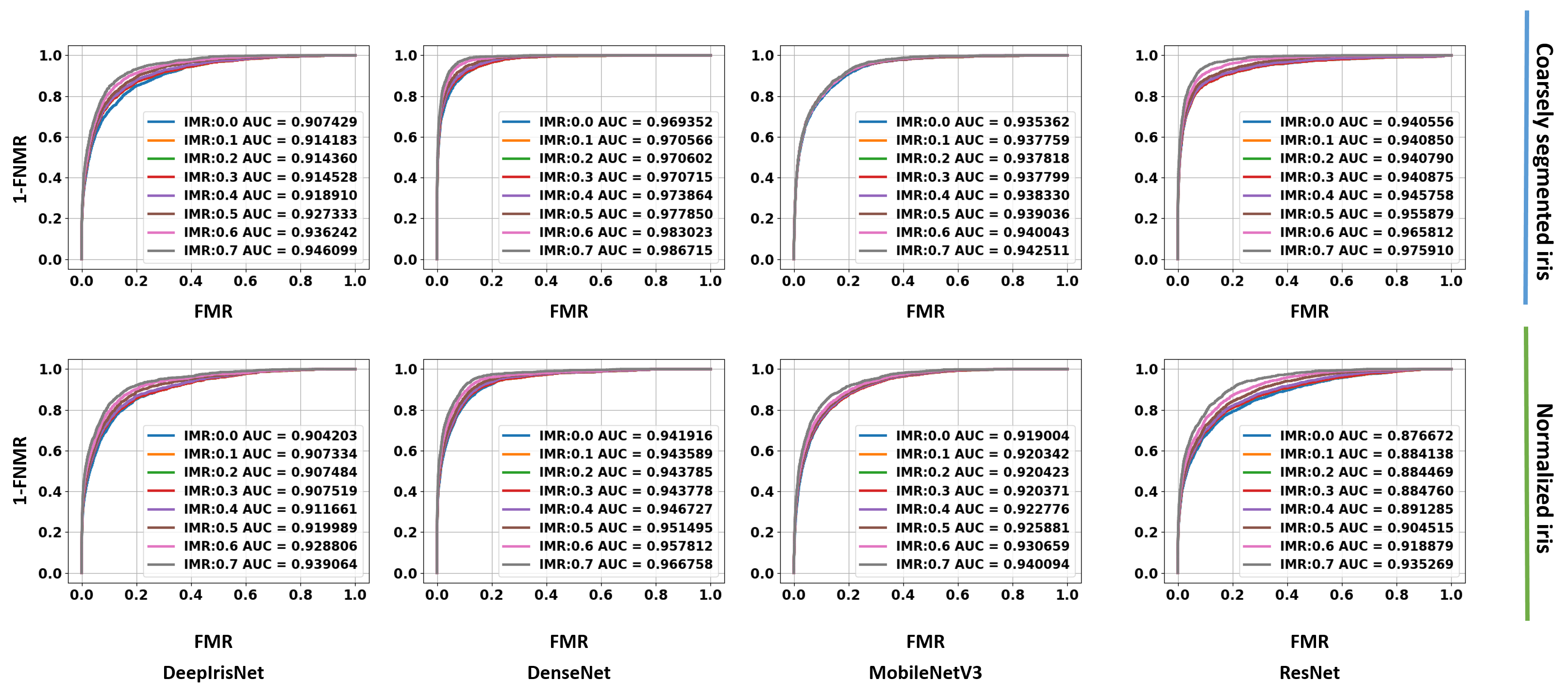}
\end{center}
  \vspace{-6mm}
   \caption{The ROC curves achieved for different deep learning settings. The plots in the first row show the evaluation results
   obtained using coarsely segmented iris region and the plots in the second show result obtained from normalized iris.
   }
\label{fig:roc_dl}
\end{figure*}

\begin{table*}[ht]
\begin{center}
\resizebox{0.92\textwidth}{!}{%
\begin{tabular}{|c|c|c|c|c|c|c|c|c|c|c|} 
\hline
\multirow{2}{*}{\begin{tabular}[c]{@{}c@{}} IMR \\ threshold \end{tabular}} & \multicolumn{2}{c|}{DCT-SHD}        & \multicolumn{2}{c|}{LG-SHD} & \multicolumn{2}{c|}{CSBCA} & \multicolumn{2}{c|}{DCT-HD} & \multicolumn{2}{c|}{LG-HD}  \\ 
\cline{2-11}
                                                                            & EER              & FMR10            & EER    & FMR10              & EER    & FMR10             & EER    & FMR10              & EER    & FMR10              \\ 
\hline
IMR 0.0                                                                     & \textbf{0.3469}  & \textbf{0.5644}  & 0.3502 & 0.5781             & 0.4032 & 0.7170            & 0.3750 & 0.6397             & 0.3663 & 0.6064             \\ 
\hline
IMR 0.7                                                                     & \textbf{0.3166}  & 0.4770           & 0.3132 & \textbf{0.4596}    & 0.3458 & 0.6619            & 0.3497 & 0.5713             & 0.3347 & 0.5313             \\
\hline
\end{tabular}}
\vspace{-2mm}
\caption{The EER and FMR10  for the different experimental settings and two IMR thresholds 0.0 and 0.7. The lowest FMR10 and EER are in bold for each IMR threshold. 
One can notice the lower errors achieved by the DCT-SHD and LG-SHD.
}
\label{tab:eer}
\vspace{-2mm}
\end{center}
\end{table*}

\begin{table*}[!h]
\centering
\resizebox{0.92\textwidth}{!}{%
\begin{tabular}{|c|c|c|c|c|c|c|c|c|c|}
\hline
\multirow{2}{*}{Modality} &
  \multirow{2}{*}{\begin{tabular}[c]{@{}c@{}
  }IMR \\ threshold\end{tabular}} &
  \multicolumn{2}{c|}{DeepIrisNet} &
  \multicolumn{2}{c|}{DenseNet} &
  \multicolumn{2}{c|}{MobileNet} &
  \multicolumn{2}{c|}{ResNet} \\ \cline{3-10} 
                                         &         & EER    & FMR10  & EER             & FMR10           & EER    & FMR10  & EER    & FMR10  \\ \hline
\multirow{2}{*}{\begin{tabular}[c]{@{}c@{}
  }Coarsely \\ segmented iris\end{tabular}} 
& IMR 0.0 & 0.1748 & 0.2728 & \textbf{0.0973} & \textbf{0.0947} & 0.1471 & 0.2144 & 0.1251 & 0.1418 \\ \cline{2-10} 
                                         & IMR 0.7 & 0.1248 & 0.1603 & \textbf{0.0580} & \textbf{0.0221} & 0.1377 & 0.1916 & 0.0774 & 0.0557 \\ \hline
\multirow{2}{*}{Normalized iris}         & IMR 0.0 & 0.1741 & 0.2786 & 0.1306          & 0.1754          & 0.1619 & 0.2492 & 0.2058 & 0.3271 \\ \cline{2-10} 
                                         & IMR 0.7 & 0.1328 & 0.1751 & 0.0945          & 0.0853          & 0.1332 & 0.1796 & 0.1472 & 0.2189 \\ \hline
\end{tabular}}
\vspace{-2mm}
\caption{The achieved verification performance of different deep learning methods evaluated on the coarsely segmented iris region and the normalized iris modality. }
\label{tab:eer_dl}
\vspace{-2mm}
\end{table*}


\vspace{-3mm}
\paragraph{Segmentation results:}
The Eye-MMS model used in this work achieved the following IoU values: a) 97.96\% on background, b) 77.85\% on sclera, c) 93.72\% on iris, d) 92.19\% on pupil, and e) 90.43\% unweighted mean IoU. 

\vspace{-3mm}
\paragraph{ Results on iris selection using proposed IMR:}
We analyze the applicability of the proposed quality metric - IMR on the verification performance by using different thresholds. We first present a histogram of the IMR values measured on the reference data and the probe data. 
Figure \ref{fig:imr}.a presents the histogram of the IMR values of the images in the reference pool. Some samples scored lower than 0.4 IMR indicating a low proportion of visible iris. When the samples with the highest IMR are selected for each unique identity, the lowest IMR value corresponds to over 0.7, as seen in Figure \ref{fig:imr}.b.  On the other hand, Figure \ref{fig:imr}.c shows the histogram of the IMR values of probe samples. One can notice that the probe samples contain some images where the iris was not visible at all, i.e. closed eyes, motivating the selection of good quality irises. The plot also shows that most probe samples had an IMR value between 0.6 and 0.9. 

Further, this IMR thresholding creates a time (sample) gap in the verification process, which is significant if the verification is performed in a continuous nature. Therefore, we analyze the amount of gap (measured by the number of images under the threshold between accepted images) in the probe consequent samples introduced by eight different IMR thresholds ( 0, 0.1, 0.2, 0.3, 0.4, 0.5, 0.6 ,and 0.7).

When samples with lower IMR values are neglected, this would produce a sample gap (SG) between consecutive frames.
Noting that the data is captured at a frame rate of 200 Hz, every single SG corresponds to 5ms of time. Having a large SG might affect the applicability to continuous authentication or, if a large SG is allowed, it will give an attacker the time frame to gain access. Therefore, it is important to study the amount and frequency of SG induced by neglecting captures with low IMR. To do that, we present a thorough analyses in Tables \ref{tab:gap}. Tables \ref{tab:gap} presents the occurrences of different SGs in each IMR thresholding setup. As noted from the Tables \ref{tab:gap}, SG 0-1 indicates the occurrences of two consecutive captures having no or one neglected capture between them, SG 2-3 indicates the occurrences of two consecutive captures having two or three neglected captures between them, and so on for SG 4-5, SG 6-7, while the SG$>$8 indicates the total number of occurrences of SG equal to 8 or more. The column indicated by "MAX SG" points out the maximum gap occurring in the setting. Table \ref{tab:gap} thus shows that increasing the IMR threshold might result in unwanted sample gaps up to 19, in the case of a 0.7 IMR threshold. However, an IMR threshold of 0.5 will only result in a few sample gaps above 6, and a max SG of 10.

\vspace{-2mm}
\paragraph{Results on iris recognition:}
The verification performances of the different evaluated deep learning approaches are presented as ROC curves
in Figure \ref{fig:roc_dl}. Beside, the EER and FMR10 values are presented in Table \ref{tab:eer} and Table \ref{tab:eer_dl} for handcraft methods, deep learning methods, respectively.
Each of the 
Figures \ref{fig:roc_dl}.a-d shows the ROCs achieved at different IMR thresholds. The ROCs achieved by the handcrafted features are not presented for space limitation as they result in much lower performance compared to deep learning methods, see Table \ref{tab:eer}. We make the following main observations from the set of experiments as noted below:
\vspace{-1.5mm}
\begin{itemize}[leftmargin=*]
   \itemsep-0.14em 
    \item One can notice that neglecting captures with low IMR enhances the performance of all the evaluated algorithms. This can be clearly explained as the iris images with high IMR present more textural information leading to accurate verification performance. The same conclusion can be made when looking at the EER and FMR10 values in Table \ref{tab:eer} and Table \ref{tab:eer_dl} where setting the IMR threshold to 0.7 reduces the  error rates consistently.
    \item The achieved verification performance by the deep learning approaches is significantly better than handcrafted features as shown in Table \ref{tab:eer} and Table \ref{tab:eer_dl} where the best verification performance is achieved by DenseNet when it is trained and evaluated on coarsely segmented iris.  
\end{itemize}
\vspace{-1.5mm}
We also make observations on the used deep learning approaches. The deep learning approaches achieved slightly better performance when  coarsely segmented iris is used in comparison to using the normalized iris as shown in Figure \ref{fig:roc_dl}. Under the most strict IMR threshold (IMR$>$0.7), the best verification performance is achieved by DenseNet model where the EER was 5.80\% when it was evaluated on coarsely segmented iris region and 9.45\% when it was evaluated on the normalized iris as shown in Table \ref{tab:eer_dl}. When no probe images are neglected, the DenseNet also achieved the best verification performance where the EER was 9.73\% for the coarsely segmented iris region and 13.06\% for the normalized iris. 
Although the DenseNet achieved the best result, the computational complexity of this model (18.5 million trainable parameters) is around 6x higher than the MobileNet (3.1 million trainable parameters).
In general, the fine-tuned models, DenseNet and ResNet, achieved slightly higher verification performance than DeepIrisNet and MobileNet trained from scratch. This result indicates that fine-tuned CNN models, originally trained for image classification, were able to capture the discriminative features of the iris image when they are fine-tuned by few training samples (8916 eye images).

As expected, the verification performances of the handcrafted approaches were lower than the deep learning ones.
It is noticeable from  the error rates in Table \ref{tab:eer} that the LG-SHD and DCT-SHD perform better than the rest of the algorithms. SHD distance achieves generally better verification performance than the HD as SHD accounts for rotational shifts. In general, the best achieved EER with handcrafted features was 34.69\% when no probes are neglected and 31.66\% when probes with IMR lower than 0.7 are neglected, considerably higher than deep learning methods.

\begin{table}[!ht]
\centering
\resizebox{\linewidth}{!}{%
\begin{tabular}{|l|l|c|c|}
\hline
\textbf{Model} & \textbf{Input size} & \textbf{No. parameters} & \textbf{Inference time} \\ \hline
DeepIrisNet-A  & 224 x 224           & 231.9m                  & 0.36s                  \\ \hline
DeepIrisNet-A  & 128 x 128           & 55.8m                   & 0.28s                  \\ \hline
DenseNet-201   & 224 x 224           & 18.5m                   & 0.34s                  \\ \hline
ResNet-50     & 224 x 224           & 23.7m                   & 0.73s                  \\ \hline
MobileNet-V3   & 224 x 224           & 3.1m                    & 0.15s                  \\ \hline
MMS-Eye        & 640 x 480           & 80k                     & 0.04s                  \\ \hline
\end{tabular}%
}
\vspace{-2mm}
\caption{Computational efficiency of deep learning approaches in this work.}
\label{tab:computational_efficiency}
\vspace{-4mm}
\end{table}

\begin{table*}[!ht]
\centering
\resizebox{0.98\textwidth}{!}{%
\begin{tabular}{ccccccccccccccc}
\cline{2-15}
                                  & \multicolumn{2}{c}{ (TV \textless TH)\%}                         & \multicolumn{2}{c}{ (TV \textgreater TH)\%}                         &                                  & \multicolumn{2}{c}{ (TV\textless{}TH)\%}                             & \multicolumn{2}{c}{(TV\textgreater{}TH)\%}                           &                                  & \multicolumn{2}{c}{(TV\textless{}TH)\%}                             & \multicolumn{2}{c}{(TV\textgreater{}TH)\%}                          \\ \hline
\multicolumn{1}{|c|}{\textbf{User-ID}} & \multicolumn{1}{c|}{Gen-IMR 0.0}   & \multicolumn{1}{c|}{Gen-IMR 0.7} & \multicolumn{1}{c|}{Imp-IMR 0.0}   & \multicolumn{1}{c|}{Imp-IMR 0.7}    & \multicolumn{1}{c|}{\textbf{User-ID}} & \multicolumn{1}{c|}{Gen-IMR 0.0}    & \multicolumn{1}{c|}{Gen-IMR 0.7}    & \multicolumn{1}{c|}{Imp-IMR0.0}     & \multicolumn{1}{c|}{Imp-IMR 0.7}    & \multicolumn{1}{c|}{\textbf{User-ID}} & \multicolumn{1}{c|}{Gen-IMR 0.0}   & \multicolumn{1}{c|}{Gen-IMR 0.7}    & \multicolumn{1}{c|}{Imp-IMR 0.0}   & \multicolumn{1}{c|}{Imp-IMR 0.7}    \\ \hline
\multicolumn{1}{|c|}{1}           & \multicolumn{1}{c|}{\textbf{0}}    & \multicolumn{1}{c|}{\textbf{0}}  & \multicolumn{1}{c|}{2.93}          & \multicolumn{1}{c|}{\textbf{2.34}}  & \multicolumn{1}{c|}{11}          & \multicolumn{1}{c|}{\textbf{0}}     & \multicolumn{1}{c|}{\textbf{0}}     & \multicolumn{1}{c|}{3.64}           & \multicolumn{1}{c|}{\textbf{0.88}}  & \multicolumn{1}{c|}{21}          & \multicolumn{1}{c|}{91.66}         & \multicolumn{1}{c|}{\textbf{81.66}} & \multicolumn{1}{c|}{\textbf{9.27}} & \multicolumn{1}{c|}{8.59}           \\ \hline
\multicolumn{1}{|c|}{2}           & \multicolumn{1}{c|}{\textbf{0}}    & \multicolumn{1}{c|}{\textbf{0}}  & \multicolumn{1}{c|}{4.69}          & \multicolumn{1}{c|}{\textbf{2.66}}  & \multicolumn{1}{c|}{12}          & \multicolumn{1}{c|}{\textbf{11.11}} & \multicolumn{1}{c|}{12.69}          & \multicolumn{1}{c|}{0.31}           & \multicolumn{1}{c|}{\textbf{0}}     & \multicolumn{1}{c|}{22}          & \multicolumn{1}{c|}{\textbf{0}}    & \multicolumn{1}{c|}{\textbf{0}}     & \multicolumn{1}{c|}{8.21}          & \multicolumn{1}{c|}{\textbf{5.65}}  \\ \hline
\multicolumn{1}{|c|}{3}           & \multicolumn{1}{c|}{\textbf{1.53}} & \multicolumn{1}{c|}{4.61}        & \multicolumn{1}{c|}{3.65}          & \multicolumn{1}{c|}{\textbf{2.81}}  & \multicolumn{1}{c|}{13}          & \multicolumn{1}{c|}{\textbf{9.47}}  & \multicolumn{1}{c|}{\textbf{9.47}}  & \multicolumn{1}{c|}{11.51}          & \multicolumn{1}{c|}{\textbf{6.20}}  & \multicolumn{1}{c|}{23}          & \multicolumn{1}{c|}{\textbf{0}}    & \multicolumn{1}{c|}{\textbf{0}}     & \multicolumn{1}{c|}{3.79}          & \multicolumn{1}{c|}{\textbf{0.51}}  \\ \hline
\multicolumn{1}{|c|}{4}           & \multicolumn{1}{c|}{\textbf{0}}    & \multicolumn{1}{c|}{\textbf{0}}  & \multicolumn{1}{c|}{\textbf{0.10}} & \multicolumn{1}{c|}{0.92}           & \multicolumn{1}{c|}{14}          & \multicolumn{1}{c|}{\textbf{0}}     & \multicolumn{1}{c|}{30.64}          & \multicolumn{1}{c|}{21.82}          & \multicolumn{1}{c|}{\textbf{15.51}} & \multicolumn{1}{c|}{24}          & \multicolumn{1}{c|}{\textbf{0}}    & \multicolumn{1}{c|}{1.56}           & \multicolumn{1}{c|}{4.74}          & \multicolumn{1}{c|}{4.74}           \\ \hline
\multicolumn{1}{|c|}{5}           & \multicolumn{1}{c|}{\textbf{0}}    & \multicolumn{1}{c|}{\textbf{0}}  & \multicolumn{1}{c|}{7.35}          & \multicolumn{1}{c|}{\textbf{5.87}}  & \multicolumn{1}{c|}{15}          & \multicolumn{1}{c|}{\textbf{0}}     & \multicolumn{1}{c|}{\textbf{0}}     & \multicolumn{1}{c|}{5.51}           & \multicolumn{1}{c|}{\textbf{0}}     & \multicolumn{1}{c|}{25}          & \multicolumn{1}{c|}{\textbf{5.55}} & \multicolumn{1}{c|}{22.22}          & \multicolumn{1}{c|}{0.25}          & \multicolumn{1}{c|}{\textbf{0}}     \\ \hline
\multicolumn{1}{|c|}{6}           & \multicolumn{1}{c|}{\textbf{40.0}} & \multicolumn{1}{c|}{65.0}        & \multicolumn{1}{c|}{0}             & \multicolumn{1}{c|}{0}              & \multicolumn{1}{c|}{16}          & \multicolumn{1}{c|}{\textbf{0}}     & \multicolumn{1}{c|}{3.94}           & \multicolumn{1}{c|}{\textbf{11.34}} & \multicolumn{1}{c|}{11.71}          & \multicolumn{1}{c|}{26}          & \multicolumn{1}{c|}{\textbf{0}}    & \multicolumn{1}{c|}{\textbf{0}}     & \multicolumn{1}{c|}{25.78}         & \multicolumn{1}{c|}{\textbf{13.78}} \\ \hline
\multicolumn{1}{|c|}{7}           & \multicolumn{1}{c|}{\textbf{0}}    & \multicolumn{1}{c|}{1.90}        & \multicolumn{1}{c|}{21.97}         & \multicolumn{1}{c|}{\textbf{10.72}} & \multicolumn{1}{c|}{17}          & \multicolumn{1}{c|}{97.89}          & \multicolumn{1}{c|}{\textbf{96.84}} & \multicolumn{1}{c|}{6.25}           & \multicolumn{1}{c|}{\textbf{5.62}}  & \multicolumn{1}{c|}{27}          & \multicolumn{1}{c|}{18.57}         & \multicolumn{1}{c|}{\textbf{5.71}}  & \multicolumn{1}{c|}{4.45}          & \multicolumn{1}{c|}{\textbf{0}}     \\ \hline
\multicolumn{1}{|c|}{8}           & \multicolumn{1}{c|}{\textbf{0}}    & \multicolumn{1}{c|}{\textbf{0}}  & \multicolumn{1}{c|}{20.27}         & \multicolumn{1}{c|}{\textbf{10.34}} & \multicolumn{1}{c|}{18}          & \multicolumn{1}{c|}{\textbf{0}}     & \multicolumn{1}{c|}{1.81}           & \multicolumn{1}{c|}{1.66}           & \multicolumn{1}{c|}{\textbf{1.19}}  & \multicolumn{1}{c|}{28}          & \multicolumn{1}{c|}{14.28}         & \multicolumn{1}{c|}{\textbf{5.49}}  & \multicolumn{1}{c|}{\textbf{0.84}} & \multicolumn{1}{c|}{0.89}           \\ \hline
\multicolumn{1}{|c|}{9}           & \multicolumn{1}{c|}{\textbf{0}}    & \multicolumn{1}{c|}{\textbf{0}}  & \multicolumn{1}{c|}{9.88}          & \multicolumn{1}{c|}{\textbf{5.34}}  & \multicolumn{1}{c|}{19}          & \multicolumn{1}{c|}{\textbf{0}}     & \multicolumn{1}{c|}{\textbf{0}}     & \multicolumn{1}{c|}{12.99}          & \multicolumn{1}{c|}{\textbf{3.24}}  & \multicolumn{1}{c|}{}            & \multicolumn{1}{c|}{}              & \multicolumn{1}{c|}{}               & \multicolumn{1}{c|}{}              & \multicolumn{1}{c|}{}               \\ \hline
\multicolumn{1}{|c|}{10}          & \multicolumn{1}{c|}{\textbf{0}}    & \multicolumn{1}{c|}{2.85}        & \multicolumn{1}{c|}{0.94}          & \multicolumn{1}{c|}{\textbf{0}}     & \multicolumn{1}{c|}{20}          & \multicolumn{1}{c|}{\textbf{0}}     & \multicolumn{1}{c|}{\textbf{0}}     & \multicolumn{1}{c|}{15.18}          & \multicolumn{1}{c|}{\textbf{3.31}}  & \multicolumn{1}{c|}{}            & \multicolumn{1}{c|}{}              & \multicolumn{1}{c|}{}               & \multicolumn{1}{c|}{}              & \multicolumn{1}{c|}{}               \\ \hline
\end{tabular}%
}
\vspace{-2mm}
\caption{The achieved result by the proposed trust model. The result is reported under two IMR threshold settings, for each identity and for genuine (Gen) and importer (Imp) user scenarios.  For each identity and for each user scenario, the text in bold indicates that the model has a better performance based on the evaluated IMR thresholds. The result is reported as the percentage of the trust values (period) where it was lower than the operational threshold (genuine user scenario) and where it was higher than the operational threshold (imposter user scenario) to all calculated trust values (full session period).  }
\label{tab:trust_model}
\vspace{-3mm}
\end{table*}

\subsection{Computational analysis }
\label{ssec:computational_efficiency}
For the sake of completion and as an indicator for future works, we present the detailed analysis of computational efficiency in Table~\ref{tab:computational_efficiency}. The computational efficiency of the deep learning approaches depends on the number of trainable parameters  and the inference latency. Table~\ref{tab:computational_efficiency} presents these factors for the used deep learning models and the segmentation model.   
All evaluations are performed using Tensorflow framework (Version 1.14) running on Linux OS with Intel(R) Xeon(R) Gold 6130 CPU 2.10GHz processor. 
The evaluation only uses single core of the processor. Each extracted features are stored as four-byte floating-point
resulting in templates of 16 kilobyte (KB), 392 KB, 367.5 KB, and 4 KB for DeepIrisNet-A, ResNet-50, DenseNet-201, and MobileNetV3, respectively.

The chosen handcraft approaches are both computationally efficient and optimal for storage purposes. Each of the feature extraction is completed within 3ms (same processor) and the templates result in 915 bytes (B), 1022 B, and 336 B for 1D Log-Gabor, DCT Coefficient iriscode and CSBCA based iriscode, respectively  when stored in lossless Portable Graphics Format (png) format. The comparison of masked iriscodes using Hamming Distance takes around 2ms while the shifted version of the same takes around 6ms.

\vspace{-1mm}
\subsection{Continuous authentication results}
\vspace{-2mm}
We evaluated the continuous authentication based on the comparison scores obtained from MobileNet based on its aptness for deployability (3.1 million of trainable parameters) suited for low computational power devices in comparison to other evaluated deep learning models. Furthermore, MobileNet achieved significantly higher verification performance than handcrafted approaches. We reported the result of the continuous authentication for two separate scenarios, genuine and imposter user scenarios, motivated by \cite{bours2012continuous,DBLP:conf/fusion/DamerMB16}. In the genuine user scenario, the trust value is calculated based on scores obtained by comparing probes with a reference of the same identity. In the imposter user scenario, the trust value is calculated based on scores obtained by comparing reference of a subject identity with probes from all other identities. 
The result is reported for two different IMR thresholds, IMR 0.0 and IMR 0.7 to account for the impact of the proposed quality metric. The $T$, threshold producing EER, was 0.5131 when the solution is evaluated using IMR 0.0 and 0.5452 when evaluated using IMR 0.7. 
Table \ref{tab:trust_model} summarizes the achieved result for each identity. The result is reported per identity for genuine user scenario as a percentage of the trust value updates where it falls below $T$ to all trust values (false rejection), i.e. the time percentage of a session where the genuine user is rejected, and for imposter user scenario as a percentage of the trust value where it is higher than the $T$ to all trust values (false acceptance), i.e. the time percentage of a session where the imposter user is accepted. Table \ref{tab:trust_model} shows that the model has very similar behaviour in the genuine user scenario when IMR threshold is set to 0 or 0.7. In the imposter user scenario, the model performed better when neglecting captures with IMR lower than 0.7 than the case when no probe is neglected.

\vspace{-2mm}
\section{Conclusions}
\label{sec:conclusions}
\vspace{-1mm}
Considering the increasing use of AR/VR technologies in novel fields and the associated developments in HMD devices, this work points out the possibility of using the in-built cameras of such devices for iris recognition.
This work takes into account the limited computational power commonly associated with such devices and evaluated the verification performance and computational efficiency of the chosen segmentation model, as well as four deep learning and three handcrafted approaches. We benchmark the verification performance of these algorithms on a realistic database captured using HMD. The overall verification result showed that the deep learning approaches reported better performance than handcrafted approaches, where the best performance was achieved by DenseNet. However, the computational cost of deep learning approaches is higher than handcrafted features ones due to the millions of trainable parameters which motivated future work on applying model compression technique such as parameter pruning to reduce the computational cost. 
We proposed a tailored quality metric for iris image selection based on the relative proportion of the visible iris, showing the effect on the verification performance. The proposed approach has shown significant improvement in verification accuracy accounting for the non-collaborative capture of the iris. As another novel component, we also presented a continuous authentication model based on MobileNet model.


\paragraph{Acknowledgment:}
This research work has been funded by the German Federal Ministry of Education and Research and the Hessen State Ministry for Higher Education, Research and the Arts within their joint support of the National Research Center for Applied Cybersecurity ATHENE.

{\small
\bibliographystyle{ieee}
\bibliography{ijcb2020_template}
}

\end{document}